\documentclass{article} 
\usepackage{nips13submit_e,times}
\usepackage{hyperref}
\usepackage{url}
\usepackage{graphicx}

\title{Attention on Attention:  Architectures for Visual Question Answering (VQA)}

\author{
Jasdeep Singh \\
Stanford University\\
\texttt{jasdeep@stanford.edu} \\
\And
Vincent Ying \\
Stanford University\\
\texttt{vhying@stanford.edu} \\
\And
Alex Nutkiewicz \\
Stanford University\\
\texttt{alexer@stanford.edu} \\
}

\nipsfinalcopy 

\begin{document}

\maketitle

\begin{abstract}
Visual Question Answering (VQA) is an increasingly popular topic in deep learning research, requiring coordination of natural language processing and computer vision modules into a single architecture. We build upon the model which placed first in the VQA Challenge by developing thirteen new attention mechanisms and introducing a simplified classifier. We performed 300 GPU hours of extensive hyperparameter and architecture searches and were able to achieve an evaluation score of 64.78\%, outperforming the existing state-of-the-art single model's validation score of 63.15\%. The code is available at \href{https://github.com/SinghJasdeep/Attention-on-Attention-for-VQA}{\textsl{\scriptsize github.com/SinghJasdeep/Attention-on-Attention-for-VQA}.}
\end{abstract}

\section{Introduction}

Visual Question Answering (VQA) is an increasingly popular topic in deep learning research as it requires coordination of several artificial intelligence-related disciplines, including Computer Vision and Natural Language Processing. Due to its growing popularity, last year (2017) a version 2 of the VQA Challenge was initiated. Due to VQA's relative complexity and need for fine grained visual and textual processing, many intricate and highly tuned architectures led performance. We chose to build upon the relatively simple model proposed by last year's winners \cite{DBLP:journals/corr/abs-1708-02711} to investigate the role of attention and ways to improve performance.

At a high level, VQA models require two forms of information: text and images. The inputs to a VQA model are images and free-form, open-ended natural language questions about the image, and the model's goal is to produce a natural language answer about the input \cite{antol2015vqa}. We use pre-trained GloVe vectors and a GRU over tokenized questions to produce question embeddings, and a Faster R-CNN to generate objects centric features from the images. This information is then passed through an attention module to create a joint embedding of the image-question and the joint embedding is then passed through a classifier to produce the final answer.

Our project aims to investigate previous methods of implementing VQA and to better understand the characteristics of more successful network architectures for this task. We build upon previous iterations of winning VQA Challenge models by developing thirteen attention mechanisms and introducing a simplified classifier to the model. We evaluate our model against other VQA implementations via an evaluation metric used in the VQA Challenge and are able to beat the 2017 VQA Challenge winners best single model scores.

\section{Related Work}

VQA has been a rapidly growing research topic since the introduction of the seminal paper by \cite{antol2015vqa}, largely because of its interdisciplinary nature. VQA problems require the model to understand a text-based question, identify elements of an image, and evaluate how these two inputs relate to one another. Much of the progress in VQA parallels developments made in other problems, such as image captioning \cite{DBLP:journals/corr/XuBKCCSZB15}
\cite{DBLP:journals/corr/VinyalsTBE16} and textual question answering
\cite{DBLP:journals/corr/KumarISBEPOGS15}\cite{DBLP:journals/corr/XiongMS16}. 

The primary method to approach VQA tasks is based on three subcomponents: creating representations for the image and question; passing these inputs through a neural network to create a co-dependent embedding; and then generating the correct natural language response. Past work by \textbf{Xiong et al.}  has investigated several improvements to the input modules of dynamic memory networks (DMN), which were originally developed for textual question answering, in order to show that a basic DMN architecture could be utilized for visual question answering. \cite{DBLP:journals/corr/XiongMS16} However, since then model architectures for visual and textual question answering have been specializing to their domains. With the models for visual question answering preferring more and more sophisticated single pass attention mechanisms. In a related paper "Show, Attend and Tell: Neural image caption generation with visual attention," \textbf{Xu et al.} introduced an attention based model that learned to describe the content of images using two different attention modules: stochastic "hard" attention and deterministic "soft" attention. \cite{DBLP:journals/corr/XuBKCCSZB15} 

The current state-of-the-art model in VQA was developed by \textbf{Teney et al.} for the 2017 VQA Challenge, in which they show how very simple, interpretable models can achieve strong performance. Their experiments show the significance of carefully designing image features, attention mechanisms (bottom-up and top-down), gated activations, and output embeddings on model performance.
\cite{DBLP:journals/corr/abs-1708-02711}

\section{Datasets}

While many large-scale datasets have been developed for the application of VQA, we decided to utilize the VQA v2.0 dataset, which contains over 200,000 images, over 1 million questions and over 11 million answers and at least three questions per image preventing the model from inferring the question without considering the input image \cite{goyal2017making}. With this data, we do the following preprocessing:
\begin{itemize}
\item Training questions and answers are tokenized and then  trimmed/padded to a maximum length of 14 words. These tokens are then represented using 300-dimentional pre-trained Wikipedia+Gigaword GloVe word embeddings \cite{pennington2014glove}.
\item Thirty six features per image are created via passing the VQA v2.0 images  through a Faster R-CNN, with bottom-up attention, as proposed by \cite{anderson2017bottom}. The Faster R-CNN detects object centric elements in the input image. This CNN is pre-trained and is held fixed during the training of the VQA model. All images are pre-converted to Faster R-CNN features for efficiency purposes.
\end{itemize}

\section{Methodology}

Our proposed model (Figure 1) derives inspiration from the winning architecture developed by \textbf{Teney et al.} for the 2017 VQA Challenge. The model implements a joint RNN/CNN for question and image embeddings, respectively. It then uses top-down attention, guided by the question embedding, on the image embeddings. The model inputs are preprocessed GloVe embeddings and Faster R-CNN feature vectors as discussed in Section 3.  
\begin{figure}[h]
\begin{center}
\includegraphics[width=1.0\textwidth]{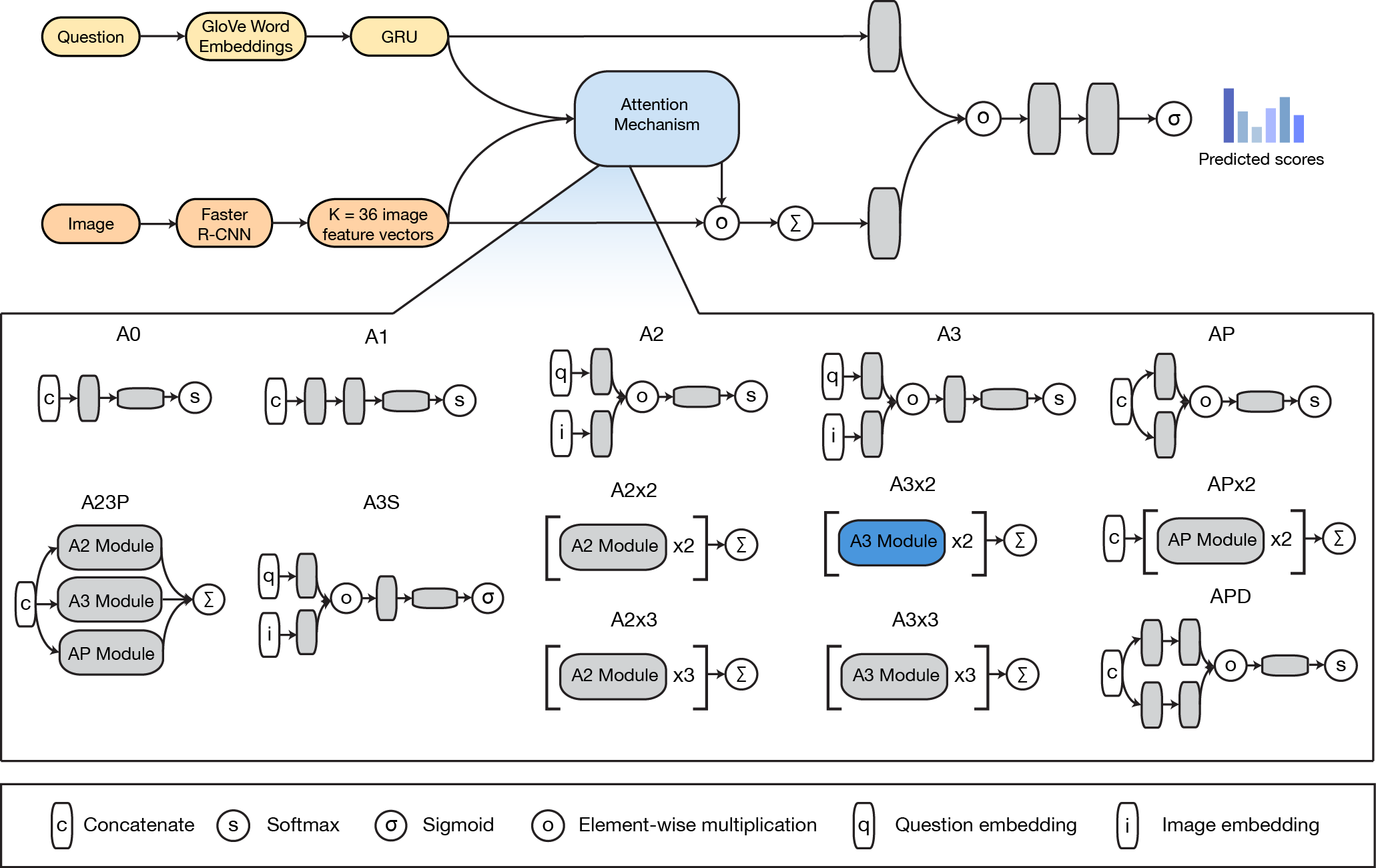}
\end{center}
\caption{Visual representation of VQA model architecture and attention modules. The specifics of each attention module are described in the text but A3x2 performed the best out of all the modules architectures investigated.}
\end{figure}
As stated in Section 3, the question inputs are tokenized and represented using GloVe word embeddings. They are then passed through a GRU to create the final question embedding. The image feature vectors along with this question embedding of size number-of-hidden-units (1280) are then passed into the dual one-way top-down attention module (A3x2). This module computes the the relevance of each of the 36 image vectors (corresponding to 36 different objects determined by the faster R-CNN) to the current question embedding using the following equations.
\begin{equation}\label{}
a_{i} =\ w f_c(f_a(\hat{i}) \circ f_b(\hat{q})) + b
\end{equation}
\begin{equation}\label{}
a_{i}' =\ w' f_{c'}(f_{a'}(\hat{i}) \circ f_{b'}(\hat{q})) + b'
\end{equation}
Where $f_x$ is a fully connected layer with a non-linearity, $w$ is a weight matrix for a linear layer with output dimension of 1, and b is a scalar.

The attention weights are then normalized using a softmax function.
\begin{equation}\label{}
	\alpha = softmax(\textbf{a}) + softmax(\textbf{a}') 
\end{equation}
The final image embedding is then created by taking a weighted sum of the original 36 image vectors using the attention scalars as weights. 
\begin{equation}\label{}
	\hat{v} = \sum_{i=1}^{K}(\alpha_i \textbf{$v_{i}$})
\end{equation}
The final image vector and the question embedding are then passed through separate one layer transformation modules, that rearrange and convert the input vectors to the same dimensions. The resulting vectors from the one layer transformation modules are then element-wise multiplied together to create the final joint embedding. This joint embedding is then given to a simple 2 layer classification module that outputs a probability via a sigmoid layer for each of the possible answers in our answer vocabulary. The word corresponding to the maximum of these output probabilities is then taken to be the predicted answer, from which the accuracy can then be calculated using the equation from \cite{DBLP:journals/corr/abs-1708-02711}:
\begin{equation}\label{}
	accuracy = \frac{1}{m_v} \ \sum_{}^{Val}(one \ hot (argmax(\hat{y})) \cdot y).
\end{equation}
Alternatively these output probabilities could also be passed to a binary cross entropy loss layer during training. 
\begin{equation}\label{}
	\mathcal{L} = \frac{1}{m_t}\sum_{i=1}^{m_t}-(y\log(\hat{y}) - (1-y)\log(1-\hat{y}))
\end{equation}
For justification of our architecture choices refer to Section 5.

\section{Experimentation}

Our initial experimentation was performed using hyperparameters identical to the ones used by \textbf{Teney et al.} However, instead of using the Adadelta optimizer we chose Adamax, and we replaced gated tanh layers with one-layer networks of twice the size because we found these modifications were able to produce a more robust model over a larger range of hyperparameters.

In our literature review of VQA models, we found one of the biggest determinants for increased model accuracy were new and improved attention mechanisms. To investigate this pattern, we implemented five new attention modules (A0, A1, A2, A3, APD), as shown in Figure 1. We evaluated these five modules, in addition to the original attention (AP) proposed by \textbf{Teney et al.} and identified the AP, A2, and A3 modules to be the most promising. The model architectures were evaluated based on their performance on the validation set (Equation 5). 

\begin{figure}[h]
\begin{center}
\includegraphics[width=1\textwidth]{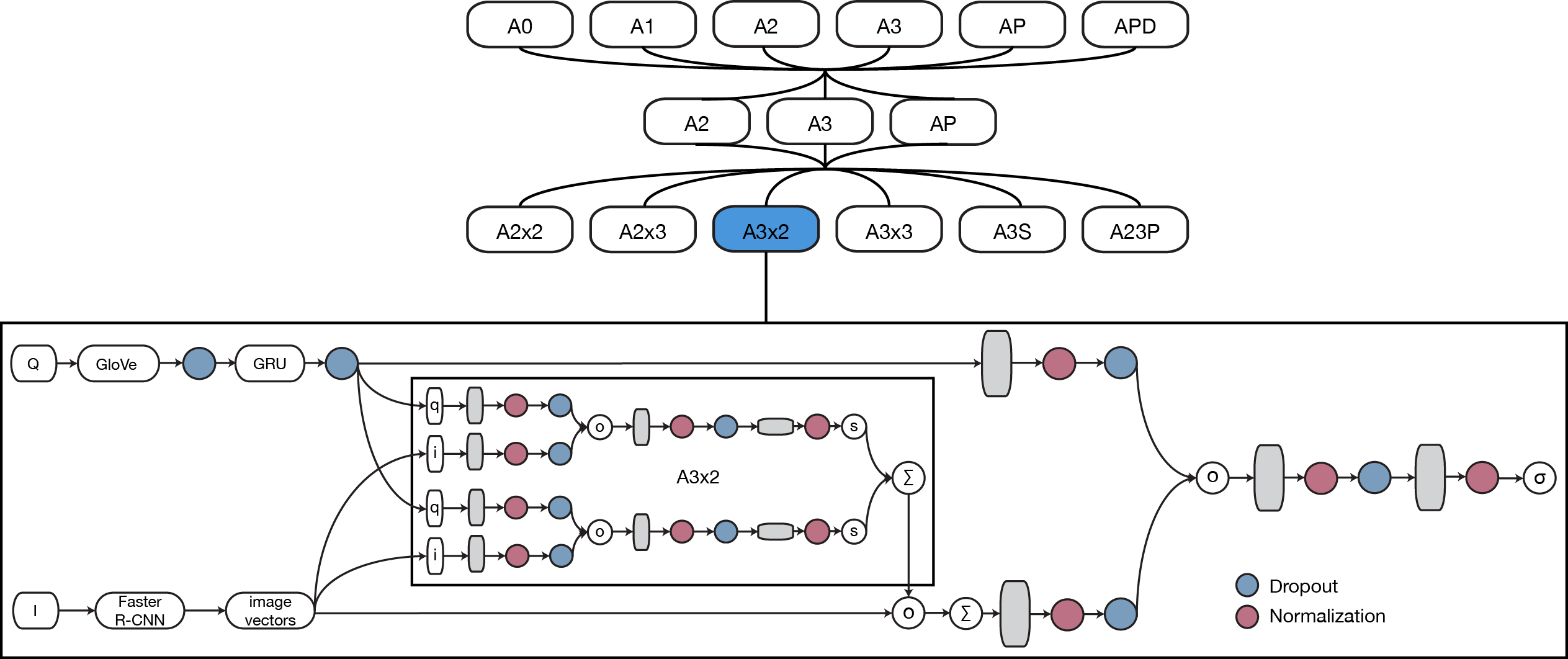}
\end{center}
\caption{Graphical representation of attention module architecture evaluation. Six primary attention modules were evaluated, and further investigation was conducted on the three optimally performing modules (A2, A3, AP). The best performing attention module, A3x2, was then used for further hyperparameter tuning.}
\end{figure}

However, we noticed that many of the attention mechanisms used in literature, including all six that we tested, had a softmax final layer. We hypothesized that this may lead to a signal bottleneck in our model and prevent the model from being able to answer questions about images that required equal attention to several regions of the image. To investigate this, we decided to add multiple attention modules to our model and also added a sigmoid final layer to our best performing attention module at the time (A3) to create A3S (Figure 1). 

After evaluating these parallel stacked attention modules and their sigmoid variants, we found the A3x2 to perform optimally and decided to pursue all further hyperparameter search using this attention module in our model. 

Hyperparameters were tuned one at a time and the general flow is presented in Figure 3. We first took our baseline model and investigated the effects of using weight normalization. We found that weight normalization (at the purple layers in Figure 2) improved performance, so we decided to keep it for further hyperparameter tuning. Next, we investigated activation functions and found the leaky ReLU to give optimal performance. At each subsequent hyperparameter step, we found the optimal value and did all following searches using that updated value. The approach may be thought of as a greedy hyperparameter search. This approach was taken over a randomized hyperparameter search due to the large space of hyperparameters searched with relatively few resources. 

\begin{figure}[h]
\begin{center}
\includegraphics[width=1\textwidth]{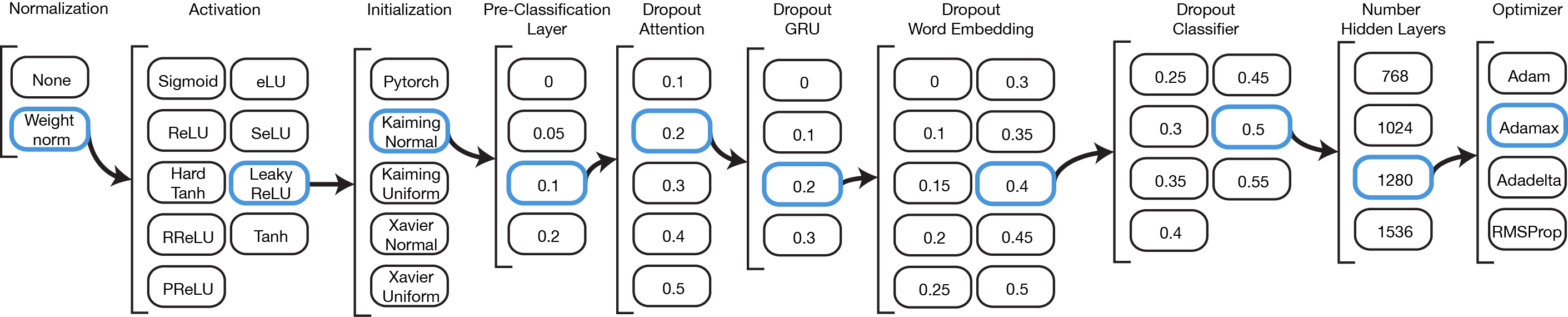}
\end{center}
\caption{Hyperparameters and selected values used for experimentation. Boxes highlighted in blue had the highest performance and were selected for the final model.}
\end{figure}

At each step, we determined the optimal hyperparameters based on validation set accuracy. However as can be seen from Figure 4 our models over-fit the training set given enough epochs. When compared to other papers, we found this to be expected because there is a large disparity between the distribution of questions in the validation and training set. This is understandable because VQA is such an open ended task, with an infinite number of possible image-question-answer triplets. 

\begin{figure}[h]
\begin{center}
\includegraphics[width=0.5\textwidth]{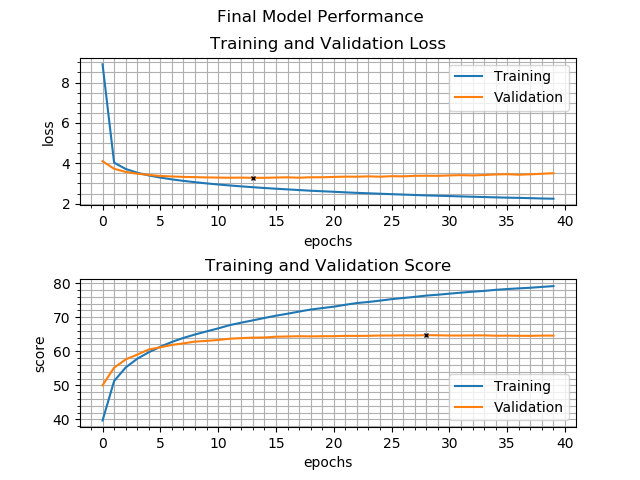}
\end{center}
\caption{Final model training and validation performance, after hyperparameter search.}
\end{figure}

Furthermore, the validation set consisted of 60,000 of the total 200,000 images while the training set consisted of a nearly similar amount of 80,000 images. Although our model may have been over-fitting the training set, it is very unlikely that dropout and activation function tuning led to over-fitting of the validation data set. 

\section{Results and Discussion}

After determining the optimal model through experimentation and tuning, we were able to achieve an evaluation score of \textbf{64.78\%}, out performing the existing state-of-the-art single model's validation score of 63.15\% (Table 1).

\begin{table}[t]
\caption{Performance of Our Model vs. State-of-the-Art}
\begin{center}
\begin{tabular}{ll}
\multicolumn{1}{c}{\bf MODEL}  &\multicolumn{1}{c}{\bf VAL PERFORMANCE SCORE}
\\ \hline \\
Our Model         &Score \textbf{64.78} \% \\
Teney et al. Model             &Score 63.15 \% \\
\end{tabular}
\end{center}
\end{table}

\begin{figure}[h]
\begin{center}
\includegraphics[width=1.0\textwidth]{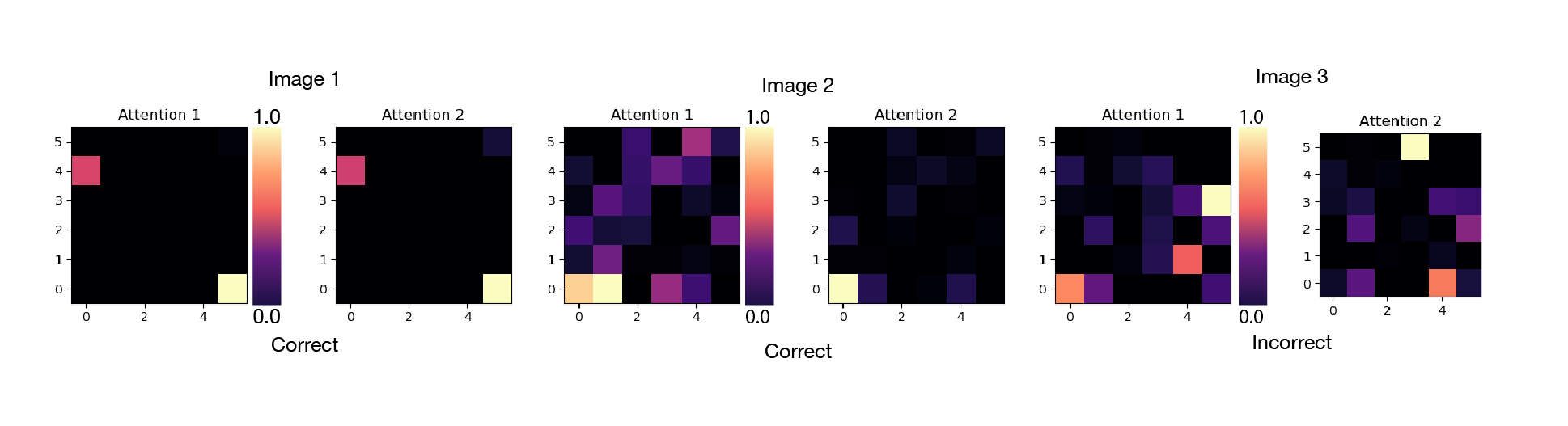}
\end{center}
\caption{Each attention module is able to pick up on different features of an input image.}
\end{figure}

We believe one of the most significant reasons our score was able to beat the state-of-the art results was because of the more sophisticated attention mechanism. The final model used attention mechanism A3x2, which takes two A3 attention mechanisms and stacks them in parallel with the ability to focus on multiple aspects of an image. Figure 5 contains three heatmaps to show how adding a second attention mechanism allows the model to learn different aspects of an input image. As you can see form Image 1, for simple attention tasks both of our  attention mechanisms are able to find the appropriate locations in the image. However, in Image 2 you can see when the task requires the need to focus highly on multiple locations in an image our model has an edge over previously presented models, which in theory leads to its increased accuracy. However for more complicated tasks such as image 3, the dual attention mechanism seems to get confused, providing no obvious advantages. 

\section{Limitations, Future Work, and Conclusion}

While our computational and time resources were limited as a result of class deadlines and budget, we were able to begin an extensive architecture and hyperparameter search. Our future work would look at the synergistic effects of some of these hyperparameters, as well as experiment with how a bi-directional attention mechanism would impact performance. Further we would like to ensemble our models so that an accurate comparison could be made with he state of the art models on the test data. However Teney et al. like many others ensembled 30 models to get state of the art performance which was unfeasible for us with our resources. 

Visual Question Answering is a unique challenge in modern Artificial Intelligence research as it combines learnings from both Computer Vision and Natural Language Processing. This paper presented our findings on what can be done to  improve performance in VQA tasks and further expands upon preexisting work by improving the model's image features, creating new attention mechanisms, and adding a simple classifier. We were able to surpass existing state-of-the art results, and we hope the insights learned from the completion of this project will inform further progress in this task.

\bibliographystyle{alpha}
\bibliography{report_bibliography}

Our code builds upon and uses all the preprocessing presented by https://github.com/hengyuan-hu/bottom-up-attention-vqa

\end{document}